\documentclass[letterpaper, 10 pt, conference]{ieeeconf}  

\usepackage[T1]{fontenc}
\usepackage[utf8]{inputenc}
\usepackage{authblk}
\usepackage{graphicx}
\usepackage[spaces,hyphens]{url}
\usepackage[linesnumbered,ruled]{algorithm2e}
\usepackage{array}
\usepackage{mathtools}
\usepackage{subcaption}
\usepackage{amssymb}
\usepackage{amsmath}
\usepackage{tikz}
\usepackage{diagbox}
\usepackage{fourier} 
\usepackage{flushend}
\usepackage{array}
\DeclareMathAlphabet{\mathcal}{OMS}{cmsy}{m}{n}
\usepackage{makecell}
\usepackage{stackengine}
\usepackage{multirow}
\usepackage[symbol]{footmisc}
\usepackage{sidecap}
\usepackage{tablefootnote}
\usepackage[flushleft]{threeparttable}
\usepackage{booktabs,caption}
\usepackage{xcolor, soul}
\usepackage{hyperref}
\sethlcolor{yellow}

\newcommand\normx[1]{\left\Vert#1\right\Vert}
\DeclareMathOperator*{\argmax}{argmax}
\DeclareMathOperator*{\argmin}{argmin}

\IEEEoverridecommandlockouts                              

\title{\LARGE \bf
6D Pose Estimation for Textureless Objects on RGB Frames using Multi-View Optimization
}

\author{Jun Yang*, Wenjie Xue$\dagger$, Sahar Ghavidel$\dagger$, and Steven L. Waslander*
\thanks{This work was supported by Epson Canada Ltd.}
\thanks{*Jun Yang and Steven L. Waslander are with University of Toronto Institute for Aerospace Studies and Robotics Institute.
        {\tt\footnotesize \{jun.yang, steven.waslander\}@robotics.utias.utoronto.ca}}
\thanks{$\dagger$Wenjie Xue and Sahar Ghavidel are with Epson Canada
        {\tt\footnotesize \{mark.xue, sahar.ghavidel\}@ea.epson.com}}
}

\begin{document}

\maketitle
\thispagestyle{empty}
\pagestyle{empty}

\begin{abstract}
6D pose estimation of textureless objects is a valuable but challenging task for many robotic applications. In this work, we propose a framework to address this challenge using only RGB images acquired from multiple viewpoints. The core idea of our approach is to decouple 6D pose estimation into a sequential two-step process, first estimating the 3D translation and then the 3D rotation of each object. This decoupled formulation first resolves the scale and depth ambiguities in single RGB images, and uses these estimates to accurately identify the object orientation in the second stage, which is greatly simplified with an accurate scale estimate. Moreover, to accommodate the multi-modal distribution present in rotation space, we develop an optimization scheme that explicitly handles object symmetries and counteracts measurement uncertainties. In comparison to the state-of-the-art multi-view approach, we demonstrate that the proposed approach achieves substantial improvements on a challenging 6D pose estimation dataset for textureless objects.
\end{abstract}

\section{INTRODUCTION}
Texture-less rigid objects occur frequently in industrial environments and are of significant interest in many robotic applications. The task of 6D pose estimation aims to detect these objects of known geometry and estimate their 6DoF (Degree of Freedom) poses, i.e., 3D translations and 3D rotations, with respect to a global coordinate frame. In robotic manipulation tasks, accurate object poses are required for path planning and grasp executions~\cite{deng2020self,wang2019densefusion,wada2020morefusion}. For robotic navigation, 6D object poses provide useful information to the robot for localization and obstacle avoidance~\cite{salas2013slam++, yang2019cubeslam, fu2021multi, merrill2022symmetry}.

Due to the lack of appearance features, historically, the problem of 6D pose estimation for textureless objects is mainly addressed with depth data~\cite{drost2010model, bui2018regression, gao20206d, gao2021cloudaae, cai2022ove6d} or RGB-D images~\cite{doumanoglou2016recovering, wang2019densefusion, he2020pvn3d, tian2020robust, saadi2021optimizing}. These approaches can achieve strong pose estimation performance when given high-quality depth data. Despite recent advances in depth acquisition technology, commodity-level depth cameras produce depth maps with low accuracy and missing data when surfaces are too glossy or dark~\cite{chai2020deep,yang2022next}, or the object is transparent~\cite{sajjan2020clear,liu2020keypose}. Hence, in the past decade, RGB-based solutions have received a lot of attention as an alternative approach~\cite{hinterstoisser2011gradient, brachmann2016uncertainty}. Due to the advancements in deep learning, some learning-based approaches have been recently shown to significantly boost the object pose estimation performance using only RGB images~\cite{kehl2017ssd, xiang2018posecnn, sundermeyer2018implicit, peng2019pvnet, hodan2020epos}. However, due to the scale, depth, and perspective ambiguities inherent to a single viewpoint, RGB-based solutions usually have low accuracy for the final estimated 6D poses.

\begin{figure*}[t]
\centering
  \includegraphics[width=0.915\linewidth]{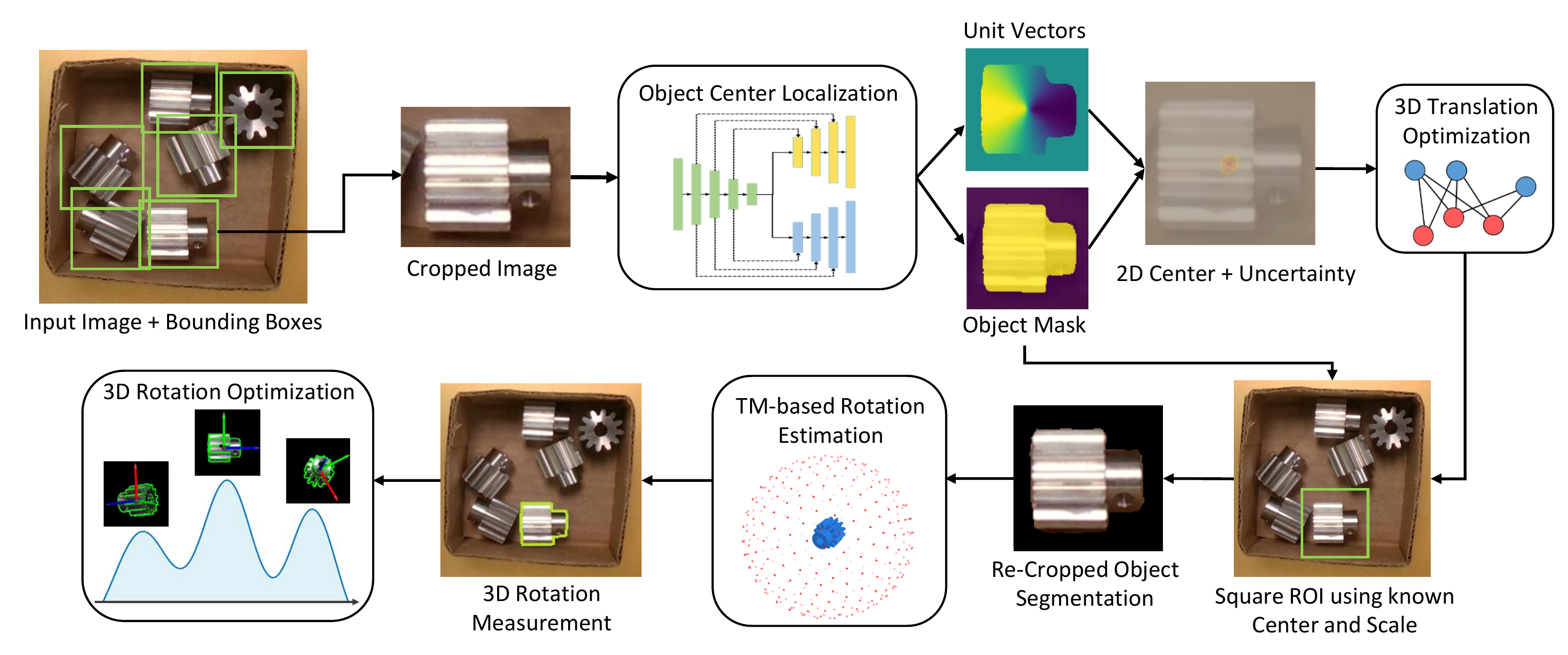}
\vspace{-0.4\baselineskip}  
\caption{An overview of the proposed multi-view object pose estimation pipeline with a two-step optimization formulation.}
\label{fig2}
\vspace{-0.95\baselineskip}
\end{figure*}

To this end, recent works utilize multiple RGB frames acquired from different viewpoints to enhance their pose estimation results~\cite{labbe2020cosypose, deng2021poserbpf, shugurov2021multi, fu2021multi, maninis2022vid2cad, merrill2022symmetry}. In particular, these approaches can be further categorized into offline batch-based solutions~\cite{labbe2020cosypose, shugurov2021multi}, where all the frames are provided at once, and incremental solutions~\cite{deng2021poserbpf, fu2021multi, maninis2022vid2cad, merrill2022symmetry}, where frames are provided sequentially. While fusing pose estimates from different viewpoints can improve the overall performance, handling extreme inconsistency, such as appearance ambiguities, rotational symmetries, and possible occlusions, is still challenging. To address these challenges, in this work, we propose a decoupled formulation to factorize the 6D pose estimation problem into a sequential two-step optimization process. Figure \ref{fig2} shows an overview of the framework. Based on the per-frame predictions of the object's segmentation mask and 2D center from neural networks, we first optimize the 3D translation and obtain the object's scale in the image. The acquired scale greatly simplifies the object rotation estimation problem with a template-matching method~\cite{hinterstoisser2011gradient}. A max-mixture formulation~\cite{olson2013inference} is finally adopted to accommodate the multi-modal output distribution present in rotation space. We conduct extensive experiments on the challenging ROBI dataset~\cite{yang2021robi}. In comparison to the state-of-the-art method CosyPose~\cite{labbe2020cosypose}, we achieve a substantial improvement with our method ($28.5\%$ and $3.4\%$ over its RGB and RGBD version, respectively).

In summary, our key contributions are:
\begin{itemize}
    \item We propose a novel 6D object pose estimation approach that decouples the problem into a sequential two-step process. This process resolves the depth ambiguities from individual RGB frames and greatly improves the estimate of rotation parameters.
    \item To deal with the multi-modal uncertainties of object rotation, we develop a rotation optimization scheme that explicitly handles the object symmetries and counteracts measurement ambiguities.
\end{itemize}

\section{RELATED WORK}
\label{sec2}
\subsection{Object Pose Estimation from a Single RGB Image}
\label{sec2a}
Many approaches have been presented in recent years to address the pose estimation problem for texture-less objects with only RGB images. Due to the lack of appearance features, traditional methods usually tackle the problem via holistic template matching techniques~\cite{hinterstoisser2011gradient,imperoli2015d,hodavn2015detection}, but are susceptible to scale change and cluttered environments. More recently, deep learning techniques, such as convolutional neural networks (CNNs), have been employed to overcome these challenges. As two pioneering methods, SSD-6D~\cite{kehl2017ssd} and PoseCNN~\cite{xiang2018posecnn} developed the CNN architectures to estimate the 6D object poses from a single RGB image. In comparison, some recent works leverage CNNs to first predict 2D object keypoints~\cite{rad2017bb8, pavlakos20176, peng2019pvnet} or dense 2D-3D correspondences~\cite{zakharov2019dpod, park2019pix2pose, hodan2020epos,haugaard2022surfemb}, and then compute the pose through 2D-3D correspondences with a PnP algorithm~\cite{lepetit2009epnp}. Although these methods show good 2D detection results, the accuracy in the final 6D poses is generally low.

\subsection{Object Pose Estimation from Multiple Viewpoints}
Multi-view approaches aim to resolve the scale and depth ambiguities that commonly occur in the single viewpoint setting and improve the accuracy of the estimated poses. Traditional works utilize local features~\cite{eidenberger2010active, collet2010efficient} and cannot handle textureless objects. Recently, the multi-view object pose estimation problem has been revisited with neural networks. These approaches used an offline, batch-based optimization formulation, where all the frames are given at once, to obtain a single consistent scene interpretation~\cite{kundu2018object, labbe2020cosypose, liu2020keypose, shugurov2021multi}. Compared to batch-based methods, other works solve the multi-view pose estimation problem in an online manner. These works estimate camera poses and object poses simultaneously, known as object-level SLAM~\cite{yang2019cubeslam, fu2021multi, wu2020eao, merrill2022symmetry}, or estimate object poses with known camera poses~\cite{deng2020self, deng2021poserbpf, maninis2022vid2cad}. Although these methods show performance improvements with only RGB images, they still face difficulty in dealing with object scales, rotational symmetries, and measurement uncertainties.

With the per-frame neural network predictions as measurements, our work resolves the depth and scale ambiguities by a decoupled formulation. It also explicitly handles rotational symmetries and measurement uncertainties within an incremental online framework.

\section{Approach Overview and Problem Formulation}
\label{sec3}
Given the 3D object model and multi-view images, the goal of 6D object pose estimation is to estimate the rigid transformation $\boldsymbol{T}_{wo} \in SE(3)$ from the object model frame ${O}$ to a global (world) frame ${W}$. We assume that we know the camera poses $\boldsymbol{T}_{wc} \in SE(3)$ with respect to the world frame. This can be done by robot forward kinematics and eye-in-hand calibration when the camera is mounted on the end-effector of a robotic arm~\cite{tsai1989new}, or off-the-shelf SLAM methods for a hand-held camera~\cite{klein2007parallel, mur2015orb}. 

Given measurements $\boldsymbol{Z}_{1:k}$ up to viewpoint $k$, we aim to estimate the posterior distribution of the 6D object pose $P\left(\boldsymbol{R}_{wo}, \boldsymbol{t}_{wo}|\boldsymbol{Z}_{1:k}\right)$. The direct computation of this distribution is generally not feasible since object translation $\boldsymbol{t}_{w,o}$ and rotation $\boldsymbol{R}_{wo}$ have distinct distributions. Specifically, the translation distribution $P\left(\boldsymbol{t}_{wo}\right)$ is straightforward and expected to be unimodal. In contrast, the distribution for object rotation $P\left(\boldsymbol{R}_{wo}\right)$ is less obvious due to complex uncertainties arising from shape symmetries, appearance ambiguities, and possible occlusions. Inspired by \cite{deng2021poserbpf}, we decouple the pose posterior $P\left(\boldsymbol{R}_{wo}, \boldsymbol{t}_{wo}|\boldsymbol{Z}_{1:k}\right)$ into:
\begin{equation}
\label{equ1}
    P\left(\boldsymbol{R}_{wo}, \boldsymbol{t}_{wo}|\boldsymbol{Z}_{1:k}\right) = P\left(\boldsymbol{R}_{wo}|\boldsymbol{Z}_{1:k}, \boldsymbol{t}_{wo}\right) P\left(\boldsymbol{t}_{wo}|\boldsymbol{Z}_{1:k}\right)
\end{equation}
where $P\left(\boldsymbol{t}_{wo}|\boldsymbol{Z}_{1:k}\right)$ can be formulated as a unimodal Gaussian distribution $\boldsymbol{\mathcal{N}\left(\boldsymbol{t}_{wo}|\mu,\Sigma\right)}$. $P\left(\boldsymbol{R}_{wo}|\boldsymbol{Z}_{1:k}, \boldsymbol{t}_{wo}\right)$ is the rotation distribution
conditioned on the input images $\boldsymbol{Z}_{1:k}$ and the 3D translation $\boldsymbol{t}_{w,o}$. To represent the complex rotation uncertainties, similar to \cite{eidenberger2010active}, we formulate $P\left(\boldsymbol{R}_{wo}|\boldsymbol{Z}_{1:k}, \boldsymbol{t}_{wo}\right)$ as the mixture of Gaussian distribution:
\begin{equation}
\label{equ2}
    P\left(\boldsymbol{R}_{wo}|\boldsymbol{Z}_{1:k}, \boldsymbol{t}_{wo}\right) = \sum_{i=1}^{N}w_i \mathcal{N}\left(\boldsymbol{R}_{wo}|\boldsymbol{\mu}_i,\boldsymbol{\Sigma}_i\right)
\end{equation}
which consists of $N$ Gaussian components. The coefficient $w_i$ denotes the weight of the mixture component. $\boldsymbol{\mu}_i$ and $\boldsymbol{\Sigma}_i$ are the mean and covariance of $i^{th}$ component, respectively.

Our proposed decoupling formulation implies a useful correlation between translation and rotation in the image domain. The 3D translation estimation $\boldsymbol{t}_{wo}$ is independent of the object's rotation and encodes the center and scale information of the object. By applying the camera pose $\boldsymbol{T}_{wc, k}$ at frame $k$, the estimated 3D translation $\boldsymbol{t}_{co, k}$ under the camera coordinate provides the scale and 2D center of the object in the image. Based on it, the per-frame object rotation measurement $\boldsymbol{R}_{co, k}$ can be estimated from its visual appearance in the image. With this formulation, our multi-view framework comprises two main steps, summarized in Figure~\ref{fig2}. In the first step (Section~\ref{Sec:translation}), we estimate the 3D translation $\boldsymbol{t}_{wo}$ by integrating the per-frame neural network outputs into an optimization formulation. The network outputs the segmentation mask and the 2D projection $\boldsymbol{u}_k$ of the object's 3D center, and the object's 3D translation $\boldsymbol{t}_{wo}$ is estimated by minimizing the 2D re-projection error across views. Given the estimated 3D translation $\boldsymbol{t}_{wo}$ and segmentation mask, in the second step (Section~\ref{Sec:rotation}), we re-crop a rotation-independent Region of Interest (RoI) for each object with the estimated scale. We then feed the RoI into a rotation estimator to get per-frame 3D rotation measurement $\boldsymbol{R}_{co, k}$. The final object rotation $\boldsymbol{R}_{wo}$ is obtained by an optimization approach with the explicitly handling of shape symmetries, and a max-mixture formulation \cite{olson2013inference} to counteract measurement uncertainties.

\begin{figure}[t]
\centering
  \includegraphics[width=\linewidth]{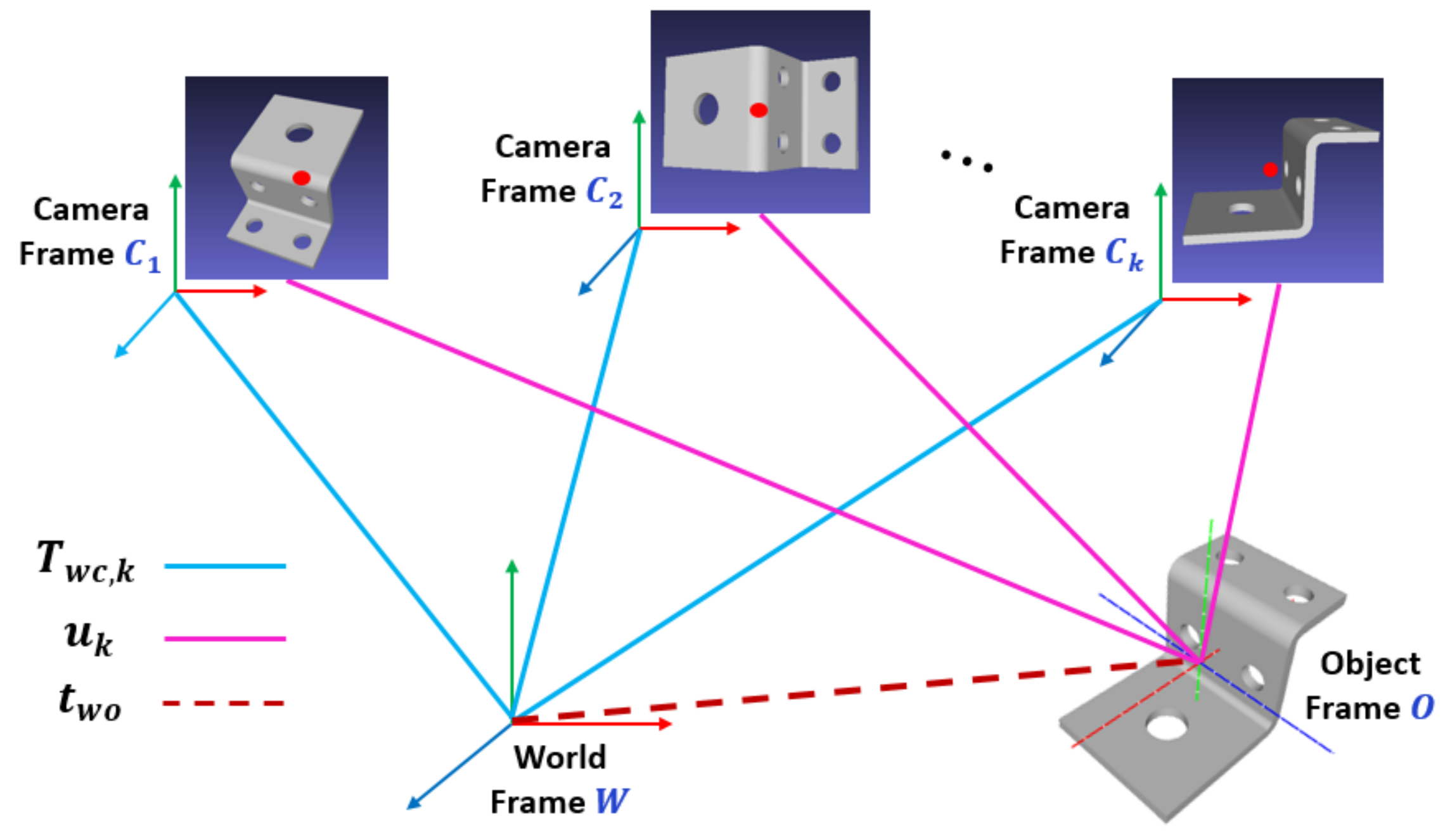}
\caption{Illustration of the object, world, and camera coordinates. The 3D translation $\boldsymbol{t}_{wo}$ is the coordinate of the object model origin in the world coordinate. We can estimate the translation by localizing the per-frame 2D center of the object, $\boldsymbol{u}_{k}$, and minimizing the re-projection errors with known camera poses $\boldsymbol{T}_{wc, k}$.}
\label{fig3}
\end{figure}

\section{3D Translation Estimation}
\label{Sec:translation}
As illustrated in Figure \ref{fig3}, the 3D translation $\boldsymbol{t}_{wo}$ is the coordinate of the object model origin in the world frame. Since the camera pose $\boldsymbol{T}_{wc}$ is known, it is equivalent to solve the translation from the object model origin to the camera optical center, $\boldsymbol{t}_{co} = {\left[t_x, t_y, t_z \right]}^T$. Given an RGB image from an arbitrary camera viewpoint, the translation $\boldsymbol{t}_{co}$ can be recovered by the following back-projection assuming a pinhole camera model,
\begin{equation}
\label{equ3}
    \begin{bmatrix}
        t_x \\
        t_y \\
        t_z
    \end{bmatrix}
    =
    \begin{bmatrix}
        \frac{u_x - c_x}{f_x} t_z \\
        \frac{u_y - c_y}{f_y} t_z \\
        t_z
    \end{bmatrix}    
\end{equation}
where $f_x$ and $f_y$ denote the camera focal lengths, and ${\left[c_x, c_y\right]}^T$ is the principal point. We define $\boldsymbol{u}={\left[u_x, u_y\right]}^T$ as the projection of the object model origin $O$ and call it the 2D center of the object in the rest of the paper. We can see that if we can localize object center $\boldsymbol{u}$ in the image and estimate the depth $t_z$, then $\boldsymbol{t}_{co}$ (or $\boldsymbol{t}_{wo}$) is solved. In our framework, we predict per-frame 2D object center $\boldsymbol{u}$ using the power of the neural network and estimate the depth $t_z$ with a multi-view optimization formulation.

Our per-frame 2D object center localization network is shown in the upper part of Figure \ref{fig2}. Our network architecture is based on PVNet~\cite{peng2019pvnet}. To deal with multiple instances in the scene, we first use off-the-shelf YOLOv5~\cite{glenn_jocher_2020_4154370} to detect 2D bounding boxes of the objects. The detections are then cropped and resized to 128x128 before being fed into the network. The network predicts pixel-wise binary labels and a 2D vector field towards the object center. A RANSAC-based voting scheme is finally utilized to estimate the mean $\boldsymbol{u}_k$ and covariance $\boldsymbol{\Sigma}_k$ of the object center at frame $k$. For more details of the object center localization prediction, we refer the reader to \cite{peng2019pvnet}.

Given a sequence of measurements, we can estimate the object 3D translation $\boldsymbol{t}_{wo}$ based on the maximum likelihood estimation (MLE) formulation. By assuming the uni-modal Gaussian error model, we solve it with a nonlinear least squares (NLLS) optimization approach. The optimization is formulated by creating measurement residuals that constrain the object translation $\boldsymbol{t}_{wo}$ with the object center $\boldsymbol{u}_k, \boldsymbol{\Sigma}_k$ and known camera pose $\boldsymbol{T}_{wc, k}$ at viewpoint $k$,
\begin{equation}
\label{equ_trans_residual}
    \boldsymbol{r}_k\left(\boldsymbol{t}_{wo}\right) = \pi\left(\boldsymbol{T}_{wc, k}^{-1}\boldsymbol{t}_{wo}\right) - \boldsymbol{u}_k
\end{equation}
where $\pi$ is the perspective projection function. The full problem becomes the minimization of the cost $L$ across all the viewpoints,
\begin{equation}
\label{equ_trans_loss}
    L = \sum_k \rho_H \left(\boldsymbol{r}_k^T \boldsymbol{\Sigma}_k^{-1} \boldsymbol{r}_k \right)
\end{equation}
where $\boldsymbol{\Sigma}_k$ is the covariance matrix estimated by the localization network for the object center $\boldsymbol{u}_k$, and $\rho_H$ is the Huber norm to reduce the impact of outliers for the optimization. We initialize each object's translation $\boldsymbol{t}_{wo}$ using the diagonal length of its 2D bounding box, similar to~\cite{kehl2017ssd}, from the first frame. With the known camera pose $\boldsymbol{T}_{wc, k}$, we perform object association based on epipolar geometry constraints and the estimated translation $\boldsymbol{t}_{wo, 1:k-1}$ up to viewpoint $k-1$. Detections that are not associated with any existing objects are initialized as new objects.

We solve the NLLS problem (Equation \ref{equ_trans_residual} and \ref{equ_trans_loss}) in an iterative Gauss-Newton procedure:
\begin{equation}
    \left( \boldsymbol{J}_{\boldsymbol{t}_{wo}}^T  \boldsymbol{\Sigma}_{\boldsymbol{z}}^{-1} \boldsymbol{J}_{\boldsymbol{t}_{wo}} \right) \delta \boldsymbol{t}_{wo} = \boldsymbol{J}_{\boldsymbol{t}_{wo}}^T \boldsymbol{\Sigma}_{\boldsymbol{z}}^{-1} \boldsymbol{r}\left(\boldsymbol{t}_{wo}\right)
\end{equation}
where $\boldsymbol{\Sigma}_{\boldsymbol{z}}$ is the stacked measurement covariance matrix up to the current frame and obtained from the object center localization network (upper part of Figure \ref{fig2}). The over Jacobian, $\boldsymbol{J}_{\boldsymbol{t}_{wo}}$, is stacked up by the per-frame Jacobian matrix $\boldsymbol{J}_{\boldsymbol{t}_{wo}, k}$.

\section{3D Rotation Estimation}
\label{Sec:rotation}
The procedure of estimating the object rotation $\boldsymbol{R}_{wo}$ is shown in the lower part of Figure~\ref{fig2}. We first adopt a template-matching (TM)-based approach, LINE-2D~\cite{hinterstoisser2011gradient}, for obtaining the per-frame rotation measurement $\boldsymbol{R}_{co, k}$. The acquired measurements from multiple viewpoints are then integrated into an optimization scheme. We handle the rotational symmetries explicitly given the object CAD model. To counteract the measurement uncertainties (e.g., from appearance ambiguities), a max-mixture formulation~\cite{olson2013inference} is used to recover a globally consistent set of object pose estimates. Note that the acquisition of the rotation measurement $\boldsymbol{R}_{co, k}$ is not limited to the LINE-2D~\cite{hinterstoisser2011gradient} or TM-based approaches and can be superseded by other holistic-based methods~\cite{liu2012fast,imperoli2015d,kehl2017ssd,sundermeyer2018implicit}.

\begin{figure}[t]
\centering
\begin{subfigure}{0.415\textwidth}
  \includegraphics[width=\linewidth]{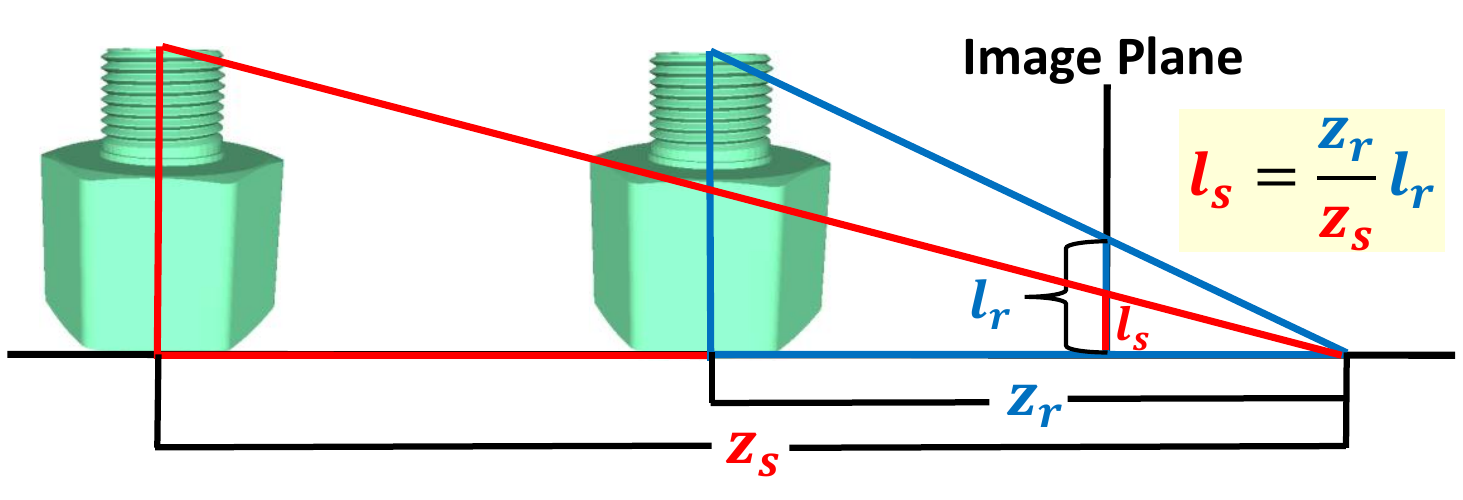}
  \vspace{-1.4\baselineskip}
  \caption{}
  \vspace{0.2\baselineskip}
  \label{fig5a}
\end{subfigure}
\begin{subfigure}{0.41\textwidth}
    \includegraphics[width=\linewidth]{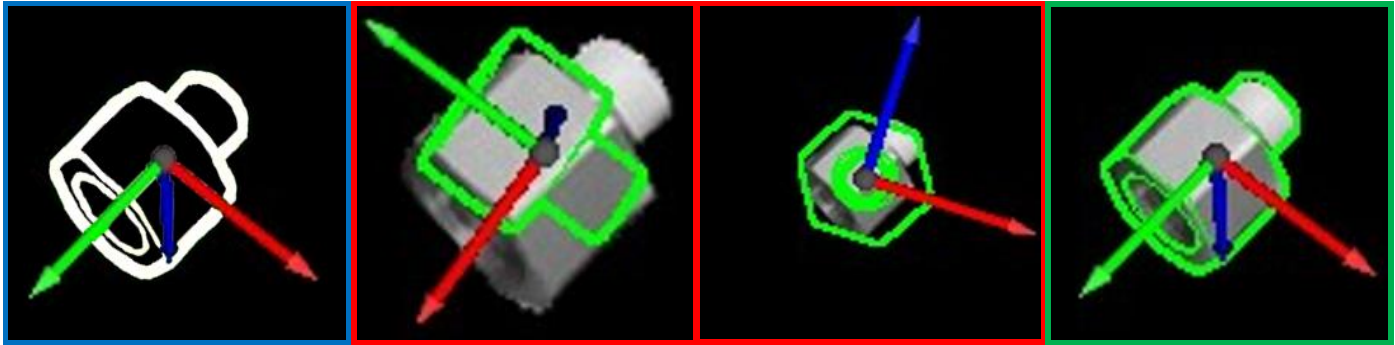}
    \vspace{-1.2\baselineskip}
    \caption{}
    \label{fig5b}
\end{subfigure}
\vspace{-0.4\baselineskip}
\caption{(a). The inference of object size $l_s$ from their projective ratio. (b) Left: the rendered object template at a canonical distance. Middle: incorrect rotation estimates due to the scale change. Right: re-cropped object RoI using the translation estimate, leading to the correct result.}
\label{fig5}
\vspace{-0.8\baselineskip}
\end{figure}

\subsection{Per-Frame Rotation Measurement}
\label{sec_rot_meas}
Given the object 3D model, LINE-2D renders object templates from a view sphere in the offline training stage (bottom middle in Figure~\ref{fig2}). At run-time, it utilizes the gradient response on input RGB or grayscale images for template matching. A confidence score is provided based on the matching quality. In general, the TM-based approach suffers from scale change issues, and the object templates need to be generated at multiple distances and scales. In our work, instead of training the multi-scale templates, which increases the run-time complexity, we fix the 3D translation to a canonical centroid distance $\boldsymbol{t}_{r} = \left[0, 0, z_r\right]$. At run-time, given the 3D translation hypothesis $\boldsymbol{t}_{co} = \left[x_s, y_s, z_s\right]$ from object origin to camera center (obtained from $\boldsymbol{t}_{wo}$ and camera pose $\boldsymbol{T}_{wc}$), we can re-crop the RoI from the image. The RoI size $l_s$ is determined by $l_s = \frac{z_r}{z_s}l_r$, where $l_r$ and $z_r$ are the RoI size and canonical distance at training time, respectively. This process is illustrated in Figure~\ref{fig5a}. Note that the RoI is a square region here and is independent of the object's rotation. To further reduce the gap between rendered templates and RoI images, we take the segmentation mask from the object center localization network (upper part of Figure~\ref{fig2}) and then feed the re-cropped object RoI into the LINE-2D estimator to get a per-frame rotation measurement $\boldsymbol{R}_{co, k}$, as shown in Figure~\ref{fig5b}.

\subsection{Optimization formulation}
Generally, estimating object 3D rotation from a sequence of measurements can also be formulated as an MLE problem:
\begin{align}
    \hat{\boldsymbol{X}} &= \argmax_{\boldsymbol{X}} \prod_k p(\boldsymbol{z}_k|\boldsymbol{X})
\end{align}
where $\boldsymbol{X}$ denotes the object 3D rotation $\boldsymbol{R}_{wo}$ to be estimated. The measurement $\boldsymbol{z}_k$ here is the object's rotation with respect to the camera coordinate $\boldsymbol{R}_{co, k}$, obtained from Section~\ref{sec_rot_meas}. The measurement model is a function of camera pose (rotation part) $\boldsymbol{R}_{wc, k}$ and object rotation $\boldsymbol{R}_{wo}$ in world frame:
\begin{align}
    h\left(\boldsymbol{R}_{wo}, \boldsymbol{R}_{wc, k}\right) = \boldsymbol{R}_{wc, k}^{-1} \boldsymbol{R}_{wo}
\end{align}

We formulate the optimization problem by creating the residual between $\boldsymbol{R}_{wo}$ and per-frame measurement $\boldsymbol{R}_{co, k}$:
\begin{align}
    \label{rot_residual}
    \boldsymbol{r}_k \left( \boldsymbol{R}_{wo} \right)&=
    \log\left({\boldsymbol{R}_{co, k} h\left(\boldsymbol{R}_{wo}, \boldsymbol{R}_{wc, k} \right)^{-1}}\right)^{\vee}
\end{align}
where $\boldsymbol{r}_k \left( \boldsymbol{R}_{wo} \right)$ is expressed by Lie algebra ${\mathfrak{so}(3)}$. To handle rotational symmetries, we consider them explicitly together with the measurement $\boldsymbol{R}_{co, k}$ in Equation \ref{rot_residual}. Generally, when an object has symmetry, there exist a set of rotations that leave the object's appearance unchanged:
\begin{align}
    \label{rot_symmetry}
    \boldsymbol{S}\left(\boldsymbol{R}_{co}\right)&=
    \Bigl\{\boldsymbol{R}^{\prime}_{co} \in SO(3)  \:\:s.t\:\: \forall \: \mathcal{G}\bigl(\boldsymbol{R}_{co}\bigl) = \mathcal{G}\bigl(\boldsymbol{R}^{\prime}_{co}\bigl) \Bigl\}
\end{align}
where $\mathcal{G}\bigl(\boldsymbol{R}_{co}\bigl)$ is the rendered image of object under rotation $\boldsymbol{R}_{co}$ (assuming the same object translation). We can update the measurement $\boldsymbol{R}_{co, k}$ in Equation \ref{rot_residual} to $\bar{\boldsymbol{R}}_{co,k}$:
\begin{align}
\label{equ6}
    \bar{\boldsymbol{R}}_{co,k}
    &= \argmin_{\boldsymbol{R}^{\prime}_{co,k} \in \boldsymbol{S}\left(\boldsymbol{R}_{co, k}\right)} \: \normx{
    \log\left({\left(\boldsymbol{R}^{\prime}_{co,k}\right) h\left(\boldsymbol{R}_{wo}, \boldsymbol{R}_{wc, k}\right)^{-1}}\right)^{\vee}}
\end{align}
where $\normx{\cdot}$ denotes the absolute angle for a 3D rotation vector $\boldsymbol{\phi}$, and $\bar{\boldsymbol{R}}_{co,k}$ is the updated rotation measurement that has the minimal loss relative to $\boldsymbol{R}_{wo}$.

\subsection{Measurement ambiguities}
Due to complex uncertainties, such uni-modal estimates are still not sufficient to adequately represent the rotation uncertainties. To this end, we now consider the sum-mixture of Gaussians as the likelihood function:
\begin{align}
\label{equ7}
    p(\bar{\boldsymbol{z}}_k|\boldsymbol{X}) = \sum_{i=1}^N w_i \mathcal{N}\left( \boldsymbol{\mu}_i, \boldsymbol{\Sigma}_i \right)
\end{align}
where $\bar{\boldsymbol{z}}_k$ is the updated measurement (using Equation~\ref{equ6}), and each $\mathcal{N} \left( \boldsymbol{\mu}_i, \boldsymbol{\Sigma}_i \right)$ represents a distinct Gaussian distribution, and $w_i$ is the weight for component $i$. The problem with a sum-mixture is that the MLE solution is no longer simple and falls outside the common NLLS optimization approaches. Instead, we consider the max-marginal and solve the problem with the following max-mixture formulation~\cite{olson2013inference}:
\begin{align}
\label{max_mix_equ}
    p(\bar{\boldsymbol{z}}_k|\boldsymbol{X}) &= \max_{i=1:N} w_i \mathcal{N}\left( \boldsymbol{\mu}_i, \boldsymbol{\Sigma}_i \right)
\end{align}

\begin{figure}[t]
\centering
\begin{subfigure}{.45\textwidth}
  \includegraphics[width=\linewidth]{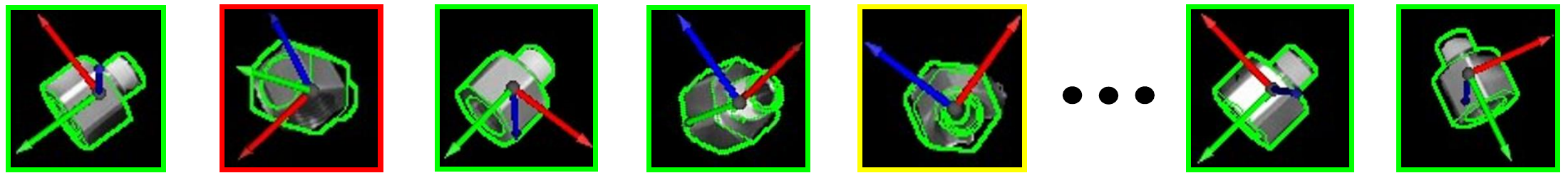}
  \caption{}
\label{fig_maxmix_a}
\end{subfigure}
\begin{subfigure}{.235\textwidth}
  \includegraphics[width=\linewidth]{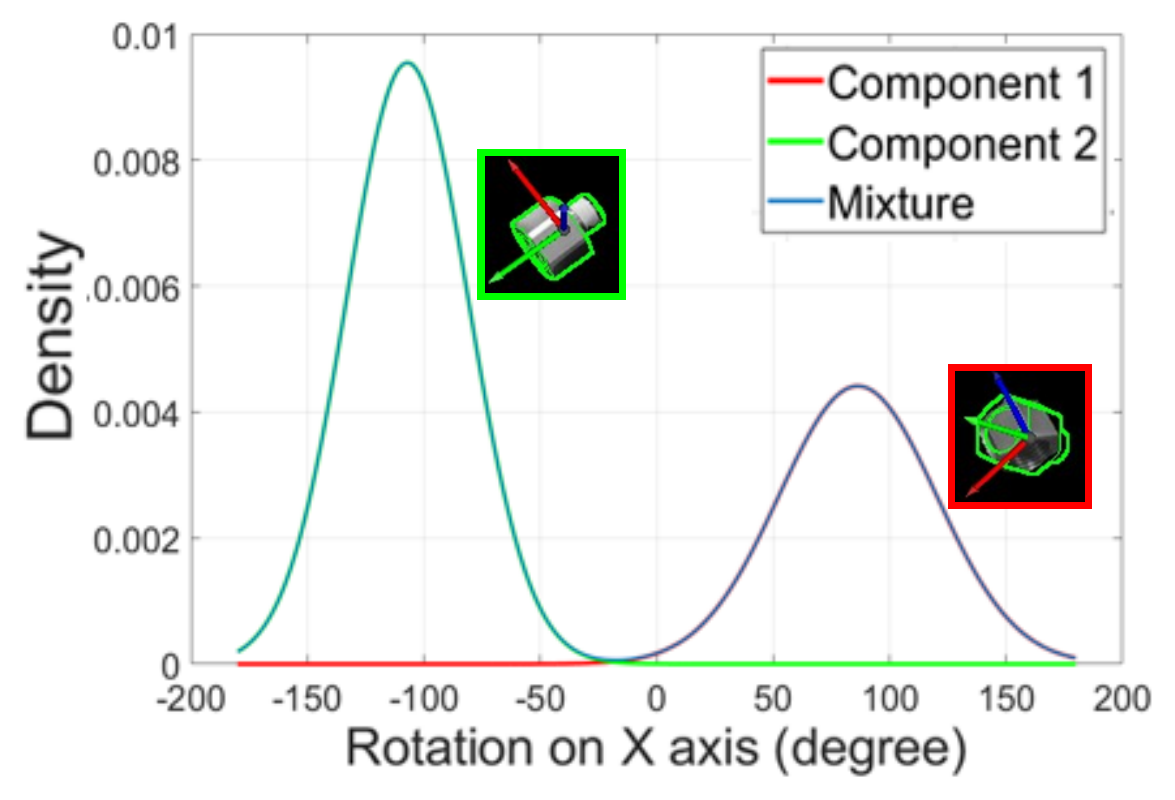}  
  \vspace{-1.4\baselineskip}
  \caption{}
  \label{fig_maxmix_b}
\end{subfigure}
\begin{subfigure}{.235\textwidth}
  \includegraphics[width=\linewidth]{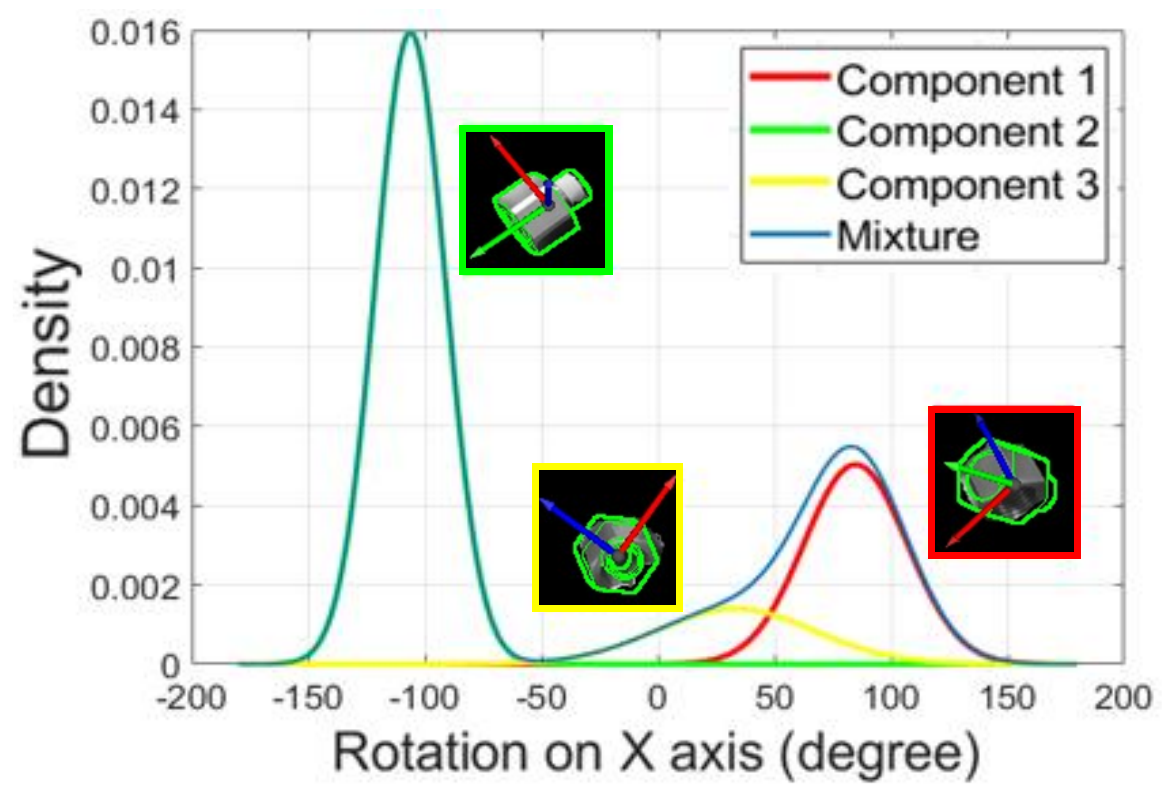}  
  \vspace{-1.4\baselineskip}
  \caption{}
  \label{fig_maxmix_c}
\end{subfigure}
\vspace{-0.6\baselineskip}
\caption{\textbf{ Max-mixtures for processing the rotation measurements.} Note that we show the distribution only on one axis for demonstration purposes. (a) Acquired rotation measurements from different viewpoints. (b) Mixture distribution after two viewpoints. (c) Mixture distribution after five viewpoints.}
\vspace{-1.2\baselineskip}
\label{fig_maxmix}
\end{figure}

The max operator acts as a selector, keeping the problem as a common NLLS optimization. Note that the Max-mixture does not make a permanent choice. In an iteration of optimization, only one of the Gaussian components is selected and optimized. In particular, given a new rotation measurement $\bar{\boldsymbol{R}}_{co,k}$ at frame $k$, we actually evaluate each Gaussian component in Equation \ref{max_mix_equ} by computing the absolute rotation angle error $\boldsymbol{\theta}_{k,i}$ between $\bar{\boldsymbol{R}}_{co,k}$ and $h\left(\boldsymbol{R}_{wo,i}, \boldsymbol{R}_{wc, k}\right)$,
\begin{align}
    \boldsymbol{\theta}_{k,i}
    &= \normx{
    \log\left(\bar{\boldsymbol{R}}_{co,k} h\left(\boldsymbol{R}_{wo}, \boldsymbol{R}_{wc, k}\right)^{-1}\right)^{\vee}}
\end{align}
and select the one with the minimal angle error. To reduce the impact of outliers, the selected Gaussian component will accept a rotation measurement only if the rotation angle error $\boldsymbol{\theta}_{k,i}$ is less than a pre-defined threshold ($30^{\circ}$ in our implementation). If the measurement $\bar{\boldsymbol{R}}_{co,k}$ is not accepted by any Gaussian component, it will be considered as a new component and added to the current Gaussian-mixture model. We optimize the object rotation $\boldsymbol{R}_{wo}$ within each component by operating on the tangent space ${\mathfrak{so}(3)}$:
\begin{equation}
\label{equ11}
    \left( \boldsymbol{J}_{\boldsymbol{\phi}_{wo}}^T  \boldsymbol{\Lambda}_{\boldsymbol{z}} \boldsymbol{J}_{\boldsymbol{\phi}_{wo}} \right) \delta \boldsymbol{\phi}_{wo} = \boldsymbol{J}_{\boldsymbol{\phi}_{wo}}^T \boldsymbol{\Lambda}_{\boldsymbol{z}} \:\:\boldsymbol{r}\left(\boldsymbol{R}_{wo}\right)
\end{equation}
where $\boldsymbol{r}\left(\boldsymbol{R}_{wo}\right)$ and $\boldsymbol{J}_{\boldsymbol{\phi}_{wo}}$ are the stacked rotation residual vector and Jacobian matrix, respectively, across multiple viewpoints. We approximate the weight matrix $\boldsymbol{\Lambda}_{\boldsymbol{z}}$ by placing the LINE-2D confidence score on its diagonal elements.

To compute the weight, $w_i$, for each Gaussian component, we accumulate the LINE-2D confidence score, $c_i$, from the rotation measurements within each individual component across the viewpoints. The weight can be approximated as: $w_i = \frac{{c}_i}{\sum_i {c}_i}$. This processing is illustrated in Figure \ref{fig_maxmix}. Given the measurements from two viewpoints, the object rotation distribution $P\left(\boldsymbol{R}_{wo}\right)$ is represented with two Gaussian components (green and red) with similar weights. The third component (yellow) is added when observing more viewpoints. As a result of receiving more rotation measurements (after five viewpoints), the weight of the correct component (green) becomes higher than the false hypotheses.

\begin{table*}[t]
\resizebox{\textwidth}{!}{
\begin{tabular}{|c|cccccc||cccccc|}
\hline
               & \multicolumn{6}{c||}{Ensenso}                                                                                                                                                         & \multicolumn{6}{c|}{RealSense}                                                                                                                                                       \\ \hline
               & \multicolumn{3}{c||}{4 Views}                                                                        & \multicolumn{3}{c||}{8 Views}                                                   & \multicolumn{3}{c||}{4 Views}                                                                        & \multicolumn{3}{c|}{8 Views}                                                   \\ \hline
Objects        & \multicolumn{2}{c|}{CosyPose}                                  & \multicolumn{1}{c||}{Ours}          & \multicolumn{2}{c|}{CosyPose}                                  & Ours          & \multicolumn{2}{c|}{CosyPose}                                  & \multicolumn{1}{c||}{Ours}          & \multicolumn{2}{c|}{CosyPose}                                  & Ours          \\ \hline
Input Modality & \multicolumn{1}{c|}{RGB}  & \multicolumn{1}{c|}{RGBD}          & \multicolumn{1}{c||}{RGB}           & \multicolumn{1}{c|}{RGB}  & \multicolumn{1}{c|}{RGBD}          & RGB           & \multicolumn{1}{c|}{RGB}  & \multicolumn{1}{c|}{RGBD}          & \multicolumn{1}{c||}{RGB}           & \multicolumn{1}{c|}{RGB}  & \multicolumn{1}{c|}{RGBD}          & RGB           \\ \hline
Tube Fitting   & \multicolumn{1}{c|}{51.6} & \multicolumn{1}{c|}{80.8}          & \multicolumn{1}{c||}{\textbf{86.1}} & \multicolumn{1}{c|}{66.2} & \multicolumn{1}{c|}{86.7}          & \textbf{88.7} & \multicolumn{1}{c|}{42.6} & \multicolumn{1}{c|}{64.7}          & \multicolumn{1}{c||}{\textbf{77.9}} & \multicolumn{1}{c|}{72.1} & \multicolumn{1}{c|}{76.5}          & \textbf{85.2} \\ \hline
Chrome Screw   & \multicolumn{1}{c|}{39.7} & \multicolumn{1}{c|}{\textbf{66.1}} & \multicolumn{1}{c||}{64.9}          & \multicolumn{1}{c|}{58.6} & \multicolumn{1}{c|}{\textbf{75.3}} & 67.8          & \multicolumn{1}{c|}{60.0} & \multicolumn{1}{c|}{62.9}          & \multicolumn{1}{c||}{\textbf{67.1}} & \multicolumn{1}{c|}{72.9} & \multicolumn{1}{c|}{\textbf{84.3}} & 81.4          \\ \hline
Eye Bolt       & \multicolumn{1}{c|}{36.5} & \multicolumn{1}{c|}{77.0}          & \multicolumn{1}{c||}{\textbf{78.4}} & \multicolumn{1}{c|}{62.1} & \multicolumn{1}{c|}{\textbf{87.8}} & 83.8          & \multicolumn{1}{c|}{26.5} & \multicolumn{1}{c|}{58.8}          & \multicolumn{1}{c||}{\textbf{79.4}} & \multicolumn{1}{c|}{55.9} & \multicolumn{1}{c|}{\textbf{91.1}} & \textbf{91.1} \\ \hline 
Gear           & \multicolumn{1}{c|}{33.3} & \multicolumn{1}{c|}{\textbf{83.9}} & \multicolumn{1}{c||}{81.5}          & \multicolumn{1}{c|}{45.6} & \multicolumn{1}{c|}{85.2}          & \textbf{86.4} & \multicolumn{1}{c|}{41.7} & \multicolumn{1}{c|}{\textbf{77.8}} & \multicolumn{1}{c||}{75.0}          & \multicolumn{1}{c|}{61.1} & \multicolumn{1}{c|}{83.3}          & \textbf{86.1} \\ \hline
Zigzag         & \multicolumn{1}{c|}{51.7} & \multicolumn{1}{c|}{75.8}          & \multicolumn{1}{c||}{\textbf{82.7}} & \multicolumn{1}{c|}{60.3} & \multicolumn{1}{c|}{89.7}          & \textbf{94.8} & \multicolumn{1}{c|}{53.6} & \multicolumn{1}{c|}{71.4}          & \multicolumn{1}{c||}{\textbf{78.6}} & \multicolumn{1}{c|}{64.3} & \multicolumn{1}{c|}{78.6}          & \textbf{89.3} \\ \hline
ALL            & \multicolumn{1}{c|}{42.6} & \multicolumn{1}{c|}{76.7}          & \multicolumn{1}{c||}{\textbf{78.7}} & \multicolumn{1}{c|}{58.6} & \multicolumn{1}{c|}{\textbf{84.9}} & 84.3          & \multicolumn{1}{c|}{44.9} & \multicolumn{1}{c|}{67.1}          & \multicolumn{1}{c||}{\textbf{75.6}} & \multicolumn{1}{c|}{65.3} & \multicolumn{1}{c|}{82.8}          & \textbf{86.6} \\ \hline
\end{tabular}}
\caption{6D object pose estimation results on Ensenso test set from ROBI dataset, evaluated with the metrics of correct detection rate. There are a total of nine scenes for the Ensenso test set and four scenes for the RealSense test set.}
 \vspace{-0.8\baselineskip}
\label{tab1}
\end{table*}

\section{EXPERIMENTS}
\label{sec4}
\subsection{Datasets, Baselines and Evaluation Metrics}
\label{sec41}
\begin{figure}[t]
\centering
\begin{subfigure}{0.43\textwidth}
  \includegraphics[width=\linewidth]{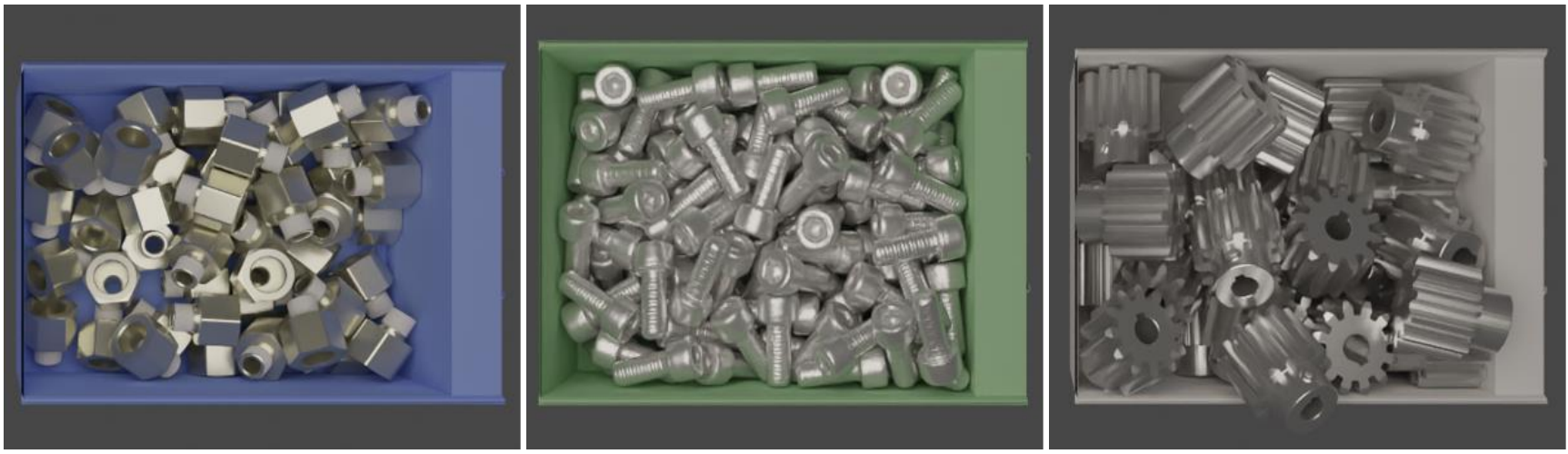}
  \vspace{-1.4\baselineskip}
  \caption{}
  \label{fig6a}
\end{subfigure}
\begin{subfigure}{0.43\textwidth}
    \includegraphics[width=\linewidth]{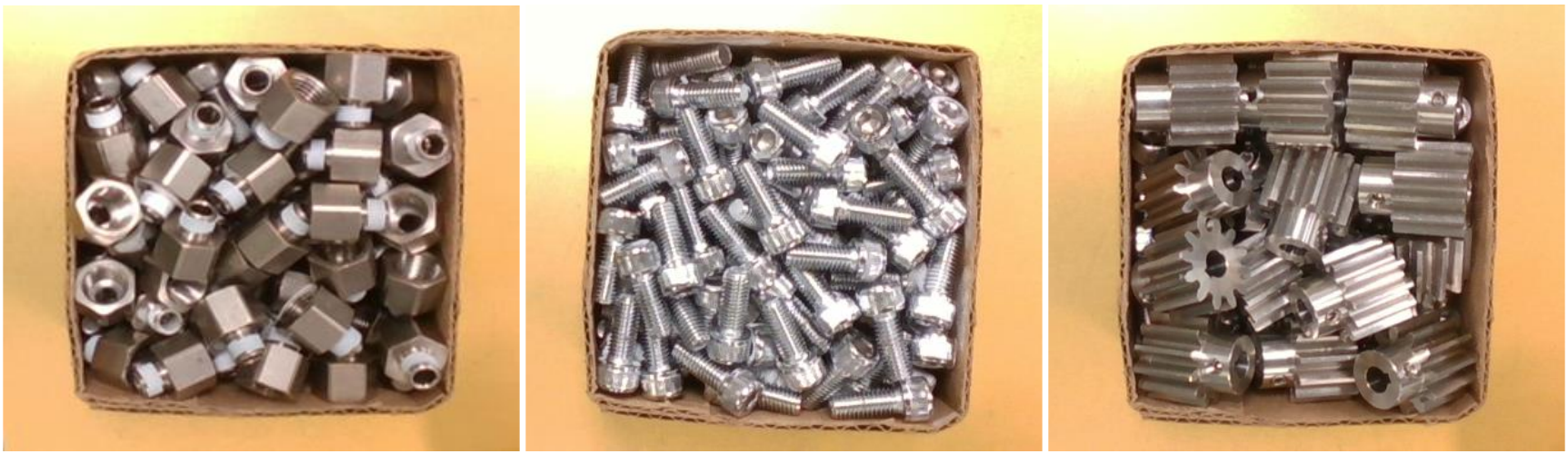}
    \vspace{-1.4\baselineskip}
      \caption{}
    \vspace{-0.4\baselineskip}      
    \label{fig6b}
\end{subfigure}
\caption{Examples of (a) the generated synthetic data and (b) real images from the RealSense Camera.}
\label{fig6}
\vspace{-1.2\baselineskip}
\end{figure}

We evaluate our framework on the recently released ROBI dataset~\cite{yang2021robi}. It provides multiple camera viewpoints and ground truth 6D poses for textureless reflective industrial parts. The objects were placed in challenging bin scenarios and captured using two sensors: a high-cost Ensenso camera and a low-cost RealSense camera. For network training purposes, we generate 80,000 synthetic images using Blender software~\cite{blender} with Bullet physics engine~\cite{coumans2016pybullet} and train our object center localization network with only synthetic data. Figure~\ref{fig6} presents some examples of our generated synthetic images. We picked five objects that are textureless and evaluated them on both Ensenso and RealSense test sets.

Quantitatively, we compare our approach with CosyPose~\cite{labbe2020cosypose}, a state-of-the-art multi-view pose fusion solution which takes the object pose estimates from individual viewpoints as the input and optimizes the overall scene consistency. Note that, CosyPose is an offline batch-based solution that is agnostic to any particular pose estimator. For a fair comparison, we use the same pose estimator (LINE-2D template matching with the same bounding boxes, object center, and segmentation mask from the object center localization network). Additionally, we provide the CosyPose with known camera poses. To feed the reliable single-view estimates to CosyPose, we use two strategies to obtain the scale information for the LINE-2D pose estimator. For the first strategy, we generate the templates at multiple distances in the training time (9 in our experiments) and perform the standard template matching in run-time. This strategy can significantly improve the single view pose estimation performance by sacrificing the run-time speed and is treated as the RGB version of CosyPose in our experiments. For the second strategy, we directly use the depth images at run-time to acquire the object scale and refer to it as the RGBD version for CosyPose. Note that, for our approach, we only use RGB images without any depth data. 

We adopt the average distance (ADD) metric~\cite{hinterstoisser2012model} for evaluation. We transform the object model points by the ground truth and the estimated 6D poses, respectively, and compute the mean of the pairwise distances between the two transformed point sets. A pose is claimed as correct if its ADD is smaller than 10\% of the object diameter. A ground truth pose will be considered only if its visibility score is larger than 75\%.

\subsection{Results}
We conduct the experiments on the ROBI dataset with a variable number of viewpoints (4 and 8). The object pose estimation results are presented in Table \ref{tab1} of five highly reflective textureless objects. The results show our method outperforms the baseline CosyPose by a wide margin on RGB data, and is competitive with the RGBD approach. On the Ensenso test set, RGBD version CosyPose achieves an overall $76.7\%$ detection with four views and $84.9\%$ with eight views. It can be considered as an upper bound of CosyPose as it uses depth data at test time. In comparison, our approach outperforms the RGB version CosyPose by a large margin of $36.1\%$ and $25.7\%$ on the 4-view and 8-view test set, respectively, and achieves the upper bound of CosyPose. On the RealSense test set, RGBD version CosyPose performs slightly worse than Ensenso results, mainly due to the poorer depth data quality from the RealSense sensor. In comparison, our approach only relies on RGB images and outperforms both RGB and RGBD versions CosyPose by $26.0\%$ and $6.1\%$, respectively.

\subsection{Ablation Study on Decoupled Formulation}
As discussed in Section~\ref{sec3} and \ref{sec_rot_meas}, the core idea of our method is the decoupling of 6D pose estimation into a sequential two-step process. This process first resolves the scale and depth ambiguities in the RGB images and greatly improves the rotation estimation performance. To justify its effectiveness, we consider an alternative version of our approach, one which simultaneously estimates the 3D translation and rotation. This version uses the same strategy to estimate the object translation. However, instead of using the provided scale from the translation estimates, it uses the multi-scale trained templates (similar to the RGB version of CosyPose) to acquire rotation measurements. Table~\ref{tab2} presents the result of our ablation study. Due to the large volume of the templates, the run-time for rotation estimation is generally slow for the simultaneous process version. In comparison, our two-step process not only operates with a much faster run-time speed but also has better overall performance.

\begin{table}[]
\resizebox{0.485\textwidth}{!}{
\begin{tabular}{|l|cc|c|}
\hline
\multirow{2}{*}{\backslashbox{Method}{Evaluation}}                                                                    & \multicolumn{2}{c|}{Detection Rate (\%)}                          & \multirow{2}{*}{\begin{tabular}[c]{@{}c@{}}Run-time\tablefootnote{We conduct the run-time analysis for \textbf{template matching only} and report with milliseconds per object. The analysis is conducted on a desktop with an Intel 3.40GHz CPU and an Nvidia RTX 2080 Ti GPU.}\\ (ms)\end{tabular}} \\ \cline{2-3}
                                                                                     & \multicolumn{1}{l|}{4 Views}       & \multicolumn{1}{l|}{8 Views} &                                                                          \\ \hline
\multicolumn{1}{|c|}{\begin{tabular}[c]{@{}c@{}}Simultaneous Process\end{tabular}} & \multicolumn{1}{c|}{74.4}          & 81.5                         & 104.5                                                                    \\ \hline
\multicolumn{1}{|c|}{\begin{tabular}[c]{@{}c@{}}Two-Step  Process\end{tabular}}    & \multicolumn{1}{c|}{\textbf{77.2}} & \textbf{85.5}                & \textbf{25.6}                                                            \\ \hline
\end{tabular}}
\caption{Ablation studies on ROBI dataset with different configurations for object pose estimation.}
 \vspace{-0.8\baselineskip}
\label{tab2}
\end{table}

\section{CONCLUSION}
\label{sec5}
In this work, we have implemented a multi-view pose estimation framework for textureless objects using only RGB images. Our core idea of our method is to decouple the posterior distribution into the 3D translation and the 3D rotation of an object and integrate the per-frame measurements with a two-step multi-view optimization formulation. This process first resolves the scale and depth ambiguities in the RGB images and greatly simplifies the per-frame rotation estimation problem. Moreover, our rotation optimization module explicitly handles the object symmetries and counteracts the measurement uncertainties with a max-mixture-based formulation. Experiments on the real ROBI dataset demonstrate the effectiveness and accuracy compared to the state-of-the-art. Future work includes joint camera pose estimation and 6D object pose estimation, and the active perception to strategically select camera viewpoints for estimating the object poses.






\bibliographystyle{ieeetr}
\bibliography{bibliography.bib}

\begin{thebibliography}{10}

\bibitem{deng2020self}
X.~Deng, Y.~Xiang, A.~Mousavian, C.~Eppner, T.~Bretl, and D.~Fox,
  ``Self-supervised 6d object pose estimation for robot manipulation,'' in {\em
  2020 IEEE International Conference on Robotics and Automation (ICRA)},
  pp.~3665--3671, IEEE, 2020.

\bibitem{wang2019densefusion}
C.~Wang, D.~Xu, Y.~Zhu, R.~Mart{\'\i}n-Mart{\'\i}n, C.~Lu, L.~Fei-Fei, and
  S.~Savarese, ``Densefusion: 6d object pose estimation by iterative dense
  fusion,'' in {\em Proceedings of the IEEE/CVF conference on computer vision
  and pattern recognition}, pp.~3343--3352, 2019.

\bibitem{wada2020morefusion}
K.~Wada, E.~Sucar, S.~James, D.~Lenton, and A.~J. Davison, ``Morefusion:
  Multi-object reasoning for 6d pose estimation from volumetric fusion,'' in
  {\em Proceedings of the IEEE/CVF conference on computer vision and pattern
  recognition}, pp.~14540--14549, 2020.

\bibitem{salas2013slam++}
R.~F. Salas-Moreno, R.~A. Newcombe, H.~Strasdat, P.~H. Kelly, and A.~J.
  Davison, ``Slam++: Simultaneous localisation and mapping at the level of
  objects,'' in {\em Proceedings of the IEEE conference on computer vision and
  pattern recognition}, pp.~1352--1359, 2013.

\bibitem{yang2019cubeslam}
S.~Yang and S.~Scherer, ``Cubeslam: Monocular 3-d object slam,'' {\em IEEE
  Transactions on Robotics}, vol.~35, no.~4, pp.~925--938, 2019.

\bibitem{fu2021multi}
J.~Fu, Q.~Huang, K.~Doherty, Y.~Wang, and J.~J. Leonard, ``A multi-hypothesis
  approach to pose ambiguity in object-based slam,'' in {\em 2021 IEEE/RSJ
  International Conference on Intelligent Robots and Systems (IROS)},
  pp.~7639--7646, IEEE, 2021.

\bibitem{merrill2022symmetry}
N.~Merrill, Y.~Guo, X.~Zuo, X.~Huang, S.~Leutenegger, X.~Peng, L.~Ren, and
  G.~Huang, ``Symmetry and uncertainty-aware object slam for 6dof object pose
  estimation,'' in {\em Proceedings of the IEEE/CVF Conference on Computer
  Vision and Pattern Recognition}, pp.~14901--14910, 2022.

\bibitem{drost2010model}
B.~Drost, M.~Ulrich, N.~Navab, and S.~Ilic, ``Model globally, match locally:
  Efficient and robust 3d object recognition,'' in {\em 2010 IEEE computer
  society conference on computer vision and pattern recognition},
  pp.~998--1005, Ieee, 2010.

\bibitem{bui2018regression}
M.~Bui, S.~Zakharov, S.~Albarqouni, S.~Ilic, and N.~Navab, ``When regression
  meets manifold learning for object recognition and pose estimation,'' in {\em
  2018 IEEE International Conference on Robotics and Automation (ICRA)},
  pp.~6140--6146, IEEE, 2018.

\bibitem{gao20206d}
G.~Gao, M.~Lauri, Y.~Wang, X.~Hu, J.~Zhang, and S.~Frintrop, ``6d object pose
  regression via supervised learning on point clouds,'' in {\em 2020 IEEE
  International Conference on Robotics and Automation (ICRA)}, pp.~3643--3649,
  IEEE, 2020.

\bibitem{gao2021cloudaae}
G.~Gao, M.~Lauri, X.~Hu, J.~Zhang, and S.~Frintrop, ``Cloudaae: Learning 6d
  object pose regression with on-line data synthesis on point clouds,'' in {\em
  2021 IEEE International Conference on Robotics and Automation (ICRA)},
  pp.~11081--11087, IEEE, 2021.

\bibitem{cai2022ove6d}
D.~Cai, J.~Heikkil{\"a}, and E.~Rahtu, ``Ove6d: Object viewpoint encoding for
  depth-based 6d object pose estimation,'' in {\em Proceedings of the IEEE/CVF
  Conference on Computer Vision and Pattern Recognition}, pp.~6803--6813, 2022.

\bibitem{doumanoglou2016recovering}
A.~Doumanoglou, R.~Kouskouridas, S.~Malassiotis, and T.-K. Kim, ``Recovering 6d
  object pose and predicting next-best-view in the crowd,'' in {\em Proceedings
  of the IEEE conference on computer vision and pattern recognition},
  pp.~3583--3592, 2016.

\bibitem{he2020pvn3d}
Y.~He, W.~Sun, H.~Huang, J.~Liu, H.~Fan, and J.~Sun, ``Pvn3d: A deep point-wise
  3d keypoints voting network for 6dof pose estimation,'' in {\em Proceedings
  of the IEEE/CVF conference on computer vision and pattern recognition},
  pp.~11632--11641, 2020.

\bibitem{tian2020robust}
M.~Tian, L.~Pan, M.~H. Ang, and G.~H. Lee, ``Robust 6d object pose estimation
  by learning rgb-d features,'' in {\em 2020 IEEE International Conference on
  Robotics and Automation (ICRA)}, pp.~6218--6224, IEEE, 2020.

\bibitem{saadi2021optimizing}
L.~Saadi, B.~Besbes, S.~Kramm, and A.~Bensrhair, ``Optimizing rgb-d fusion for
  accurate 6dof pose estimation,'' {\em IEEE Robotics and Automation Letters},
  vol.~6, no.~2, pp.~2413--2420, 2021.

\bibitem{chai2020deep}
C.-Y. Chai, Y.-P. Wu, and S.-L. Tsao, ``Deep depth fusion for black,
  transparent, reflective and texture-less objects,'' in {\em 2020 IEEE
  International Conference on Robotics and Automation (ICRA)}, pp.~6766--6772,
  IEEE, 2020.

\bibitem{yang2022next}
J.~Yang and S.~L. Waslander, ``Next-best-view prediction for active stereo
  cameras and highly reflective objects,'' in {\em 2022 International
  Conference on Robotics and Automation (ICRA)}, pp.~3684--3690, IEEE, 2022.

\bibitem{sajjan2020clear}
S.~Sajjan, M.~Moore, M.~Pan, G.~Nagaraja, J.~Lee, A.~Zeng, and S.~Song, ``Clear
  grasp: 3d shape estimation of transparent objects for manipulation,'' in {\em
  2020 IEEE International Conference on Robotics and Automation (ICRA)},
  pp.~3634--3642, IEEE, 2020.

\bibitem{liu2020keypose}
X.~Liu, R.~Jonschkowski, A.~Angelova, and K.~Konolige, ``Keypose: Multi-view 3d
  labeling and keypoint estimation for transparent objects,'' in {\em
  Proceedings of the IEEE/CVF conference on computer vision and pattern
  recognition}, pp.~11602--11610, 2020.

\bibitem{hinterstoisser2011gradient}
S.~Hinterstoisser, C.~Cagniart, S.~Ilic, P.~Sturm, N.~Navab, P.~Fua, and
  V.~Lepetit, ``Gradient response maps for real-time detection of textureless
  objects,'' {\em IEEE transactions on pattern analysis and machine
  intelligence}, vol.~34, no.~5, pp.~876--888, 2011.

\bibitem{brachmann2016uncertainty}
E.~Brachmann, F.~Michel, A.~Krull, M.~Y. Yang, S.~Gumhold, {\em et~al.},
  ``Uncertainty-driven 6d pose estimation of objects and scenes from a single
  rgb image,'' in {\em Proceedings of the IEEE conference on computer vision
  and pattern recognition}, pp.~3364--3372, 2016.

\bibitem{kehl2017ssd}
W.~Kehl, F.~Manhardt, F.~Tombari, S.~Ilic, and N.~Navab, ``Ssd-6d: Making
  rgb-based 3d detection and 6d pose estimation great again,'' in {\em
  Proceedings of the IEEE international conference on computer vision},
  pp.~1521--1529, 2017.

\bibitem{xiang2018posecnn}
Y.~Xiang, T.~Schmidt, V.~Narayanan, and D.~Fox, ``Posecnn: A convolutional
  neural network for 6d object pose estimation in cluttered scenes,'' in {\em
  Robotics: Science and Systems (RSS)}, 2018.

\bibitem{sundermeyer2018implicit}
M.~Sundermeyer, Z.-C. Marton, M.~Durner, M.~Brucker, and R.~Triebel, ``Implicit
  3d orientation learning for 6d object detection from rgb images,'' in {\em
  Proceedings of the european conference on computer vision (ECCV)},
  pp.~699--715, 2018.

\bibitem{peng2019pvnet}
S.~Peng, Y.~Liu, Q.~Huang, X.~Zhou, and H.~Bao, ``Pvnet: Pixel-wise voting
  network for 6dof pose estimation,'' in {\em Proceedings of the IEEE/CVF
  Conference on Computer Vision and Pattern Recognition}, pp.~4561--4570, 2019.

\bibitem{hodan2020epos}
T.~Hodan, D.~Barath, and J.~Matas, ``Epos: Estimating 6d pose of objects with
  symmetries,'' in {\em Proceedings of the IEEE/CVF conference on computer
  vision and pattern recognition}, pp.~11703--11712, 2020.

\bibitem{labbe2020cosypose}
Y.~Labb{\'e}, J.~Carpentier, M.~Aubry, and J.~Sivic, ``Cosypose: Consistent
  multi-view multi-object 6d pose estimation,'' in {\em European Conference on
  Computer Vision}, pp.~574--591, Springer, 2020.

\bibitem{deng2021poserbpf}
X.~Deng, A.~Mousavian, Y.~Xiang, F.~Xia, T.~Bretl, and D.~Fox, ``Poserbpf: A
  rao--blackwellized particle filter for 6-d object pose tracking,'' {\em IEEE
  Transactions on Robotics}, vol.~37, no.~5, pp.~1328--1342, 2021.

\bibitem{shugurov2021multi}
I.~Shugurov, I.~Pavlov, S.~Zakharov, and S.~Ilic, ``Multi-view object pose
  refinement with differentiable renderer,'' {\em IEEE Robotics and Automation
  Letters}, vol.~6, no.~2, pp.~2579--2586, 2021.

\bibitem{maninis2022vid2cad}
K.-K. Maninis, S.~Popov, M.~Niesser, and V.~Ferrari, ``Vid2cad: Cad model
  alignment using multi-view constraints from videos,'' {\em IEEE Transactions
  on Pattern Analysis and Machine Intelligence}, 2022.

\bibitem{olson2013inference}
E.~Olson and P.~Agarwal, ``Inference on networks of mixtures for robust robot
  mapping,'' {\em The International Journal of Robotics Research}, vol.~32,
  no.~7, pp.~826--840, 2013.

\bibitem{yang2021robi}
J.~Yang, Y.~Gao, D.~Li, and S.~L. Waslander, ``Robi: A multi-view dataset for
  reflective objects in robotic bin-picking,'' in {\em 2021 IEEE/RSJ
  International Conference on Intelligent Robots and Systems (IROS)},
  pp.~9788--9795, IEEE, 2021.

\bibitem{imperoli2015d}
M.~Imperoli and A.~Pretto, ``D co: Fast and robust registration of 3d
  textureless objects using the directional chamfer distance,'' in {\em
  Computer Vision Systems: 10th International Conference, ICVS 2015,
  Copenhagen, Denmark, July 6-9, 2015, Proceedings}, pp.~316--328, Springer,
  2015.

\bibitem{hodavn2015detection}
T.~Hoda{\v{n}}, X.~Zabulis, M.~Lourakis, {\v{S}}.~Obdr{\v{z}}{\'a}lek, and
  J.~Matas, ``Detection and fine 3d pose estimation of texture-less objects in
  rgb-d images,'' in {\em 2015 IEEE/RSJ International Conference on Intelligent
  Robots and Systems (IROS)}, pp.~4421--4428, IEEE, 2015.

\bibitem{rad2017bb8}
M.~Rad and V.~Lepetit, ``Bb8: A scalable, accurate, robust to partial occlusion
  method for predicting the 3d poses of challenging objects without using
  depth,'' in {\em Proceedings of the IEEE international conference on computer
  vision}, pp.~3828--3836, 2017.

\bibitem{pavlakos20176}
G.~Pavlakos, X.~Zhou, A.~Chan, K.~G. Derpanis, and K.~Daniilidis, ``6-dof
  object pose from semantic keypoints,'' in {\em 2017 IEEE international
  conference on robotics and automation (ICRA)}, pp.~2011--2018, IEEE, 2017.

\bibitem{zakharov2019dpod}
S.~Zakharov, I.~Shugurov, and S.~Ilic, ``Dpod: 6d pose object detector and
  refiner,'' in {\em Proceedings of the IEEE/CVF international conference on
  computer vision}, pp.~1941--1950, 2019.

\bibitem{park2019pix2pose}
K.~Park, T.~Patten, and M.~Vincze, ``Pix2pose: Pixel-wise coordinate regression
  of objects for 6d pose estimation,'' in {\em Proceedings of the IEEE/CVF
  International Conference on Computer Vision}, pp.~7668--7677, 2019.

\bibitem{haugaard2022surfemb}
R.~L. Haugaard and A.~G. Buch, ``Surfemb: Dense and continuous correspondence
  distributions for object pose estimation with learnt surface embeddings,'' in
  {\em Proceedings of the IEEE/CVF Conference on Computer Vision and Pattern
  Recognition}, pp.~6749--6758, 2022.

\bibitem{lepetit2009epnp}
V.~Lepetit, F.~Moreno-Noguer, and P.~Fua, ``Epnp: An accurate o (n) solution to
  the pnp problem,'' {\em International journal of computer vision}, vol.~81,
  no.~2, pp.~155--166, 2009.

\bibitem{eidenberger2010active}
R.~Eidenberger and J.~Scharinger, ``Active perception and scene modeling by
  planning with probabilistic 6d object poses,'' in {\em 2010 IEEE/RSJ
  International Conference on Intelligent Robots and Systems}, pp.~1036--1043,
  IEEE, 2010.

\bibitem{collet2010efficient}
A.~Collet and S.~S. Srinivasa, ``Efficient multi-view object recognition and
  full pose estimation,'' in {\em 2010 IEEE International Conference on
  Robotics and Automation}, pp.~2050--2055, IEEE, 2010.

\bibitem{kundu2018object}
J.~N. Kundu, M.~Rahul, A.~Ganeshan, and R.~V. Babu, ``Object pose estimation
  from monocular image using multi-view keypoint correspondence,'' in {\em
  European Conference on Computer Vision}, pp.~298--313, Springer, 2018.

\bibitem{wu2020eao}
Y.~Wu, Y.~Zhang, D.~Zhu, Y.~Feng, S.~Coleman, and D.~Kerr, ``Eao-slam:
  Monocular semi-dense object slam based on ensemble data association,'' in
  {\em 2020 IEEE/RSJ International Conference on Intelligent Robots and Systems
  (IROS)}, pp.~4966--4973, IEEE, 2020.

\bibitem{tsai1989new}
R.~Y. Tsai, R.~K. Lenz, {\em et~al.}, ``A new technique for fully autonomous
  and efficient 3 d robotics hand/eye calibration,'' {\em IEEE Transactions on
  robotics and automation}, vol.~5, no.~3, pp.~345--358, 1989.

\bibitem{klein2007parallel}
G.~Klein and D.~Murray, ``Parallel tracking and mapping for small ar
  workspaces,'' in {\em 2007 6th IEEE and ACM international symposium on mixed
  and augmented reality}, pp.~225--234, IEEE, 2007.

\bibitem{mur2015orb}
R.~Mur-Artal, J.~M.~M. Montiel, and J.~D. Tardos, ``Orb-slam: a versatile and
  accurate monocular slam system,'' {\em IEEE transactions on robotics},
  vol.~31, no.~5, pp.~1147--1163, 2015.

\bibitem{glenn_jocher_2020_4154370}
G.~Jocher, ``{ultralytics/yolov5: v3.1 - Bug Fixes and Performance
  Improvements}.'' \url{https://github.com/ultralytics/yolov5}, Oct. 2020.

\bibitem{liu2012fast}
M.-Y. Liu, O.~Tuzel, A.~Veeraraghavan, Y.~Taguchi, T.~K. Marks, and
  R.~Chellappa, ``Fast object localization and pose estimation in heavy clutter
  for robotic bin picking,'' {\em The International Journal of Robotics
  Research}, vol.~31, no.~8, pp.~951--973, 2012.

\bibitem{blender}
B.~O. Community, {\em Blender - a 3D modelling and rendering package}.
\newblock Blender Foundation, Stichting Blender Foundation, Amsterdam, 2018.

\bibitem{coumans2016pybullet}
E.~Coumans and Y.~Bai, ``Pybullet, a python module for physics simulation for
  games, robotics and machine learning,'' 2016.

\bibitem{hinterstoisser2012model}
S.~Hinterstoisser, V.~Lepetit, S.~Ilic, S.~Holzer, G.~Bradski, K.~Konolige, and
  N.~Navab, ``Model based training, detection and pose estimation of
  texture-less 3d objects in heavily cluttered scenes,'' in {\em Asian
  conference on computer vision}, pp.~548--562, Springer, 2012.

\end{thebibliography}

\end{document}